%% file: paper.tex
\documentclass{article}

\usepackage{arxiv}

\usepackage[utf8]{inputenc} 
\usepackage[T1]{fontenc}    

\usepackage{microtype}
\usepackage{graphicx}
\usepackage{subcaption}
\usepackage{multirow}
\usepackage{booktabs} 

\usepackage[colorlinks=true,urlcolor=black]{hyperref}
\usepackage{amsmath}
\usepackage{multirow}
\usepackage{amsfonts}
\usepackage{graphicx}
\usepackage[numbers]{natbib}
\usepackage{algorithm,algorithmic}

\usepackage{adjustbox,booktabs,colortbl}

\usepackage{cleveref}
\usepackage[title]{appendix}

\newcommand{\etal}{et al.~}
\newcommand{\ordinal}[1]{{#1}^\mathrm{th}}
\newcommand{\realset}{\mathbb{R}}
\newcommand{\tx}{\mathbf{x}}

\newcommand{\tm}{\mathbf{m}}

\newcommand{\tmin}{\mathrm{min}}
\newcommand{\tmax}{\mathrm{max}}
\newcommand{\high}{\mathrm{high}}
\newcommand{\low}{\mathrm{low}}
\newcommand{\up}{\mathrm{up}}

\makeatletter
\newcommand{\printfnsymbol}[1]{%
  \textsuperscript{\@fnsymbol{#1}}%
}
\makeatother

\title{Towards Certifiable Adversarial Sample Detection}

\author{
Ilia Shumailov \printfnsymbol{1}\\
University of Cambridge\\
{\tt\small ilia.shumailov@cl.cam.ac.uk}
\And
Yiren Zhao  \thanks{Equal Contribution}\\
University of Cambridge \\
{\tt\small yiren.zhao@cl.cam.ac.uk}
\And
Robert Mullins\\
University of Cambridge\\
{\tt\small robert.mullins@cl.cam.ac.uk}
\And
Ross Anderson\\
University of Cambridge\\
{\tt\small ross.anderson@cl.cam.ac.uk}
}

\begin{document}

\maketitle

\begin{abstract}
    Convolutional Neural Networks (CNNs) are deployed in more and more
    classification systems, but adversarial samples can be maliciously crafted to trick them, and are becoming a real threat.
    There have been various proposals to improve CNNs' adversarial robustness but these all suffer performance penalties or other limitations.
    In this paper, we provide a new approach in the form of
    a certifiable adversarial detection scheme, the Certifiable Taboo Trap (CTT).
    The system can
    provide certifiable guarantees of detection of
    adversarial inputs for certain $l_{\infty}$ sizes on a reasonable assumption, namely that the training
    data have the same distribution as the test data.
    We develop and evaluate several versions of CTT with a range of defense capabilities, training overheads
    and certifiability on adversarial samples.
    Against adversaries with various $l_p$ norms,
    CTT outperforms existing defense methods
    that focus purely on improving network robustness.
    We show that CTT has
    small false positive rates on clean test data,
    minimal compute overheads when deployed, and
    can support complex security policies.
\end{abstract}

\input{sections/intro.tex}
\input{sections/related.tex}
\input{sections/method.tex}

\input{sections/evaluation.tex}
\input{sections/conclusion.tex}
\input{sections/appendix.tex}

\bibliography{references}
\bibliographystyle{icml2020}

\end{document}

%% file: sections/intro.tex
\section{Introduction}
Convolutional Neural Networks (CNNs) give the best performance on visual applications
\citep{badrinarayanan2015segnet,krizhevsky2012imagenet,ren2015faster} and
are now spreading into safety-critical fields,
including autonomous vehicles \citep{fang2003road},
face recognition \citep{schroff2015facenet} and human action recognition \citep{ji20133d}.
However,
small perturbations can be crafted to trigger misclassifications that are not perceptible by humans \citep{goodfellow2014explaining}.
Researchers have demonstrated
adversarial samples that can
exploit face-recognition systems to break into smartphones \citep{Carlini2016}
and misdirect autonomous vehicles by perturbing road signs \citep{eykholt2018robust}.
These adversarial samples can be surprisingly portable.
Samples generated from one classifier transfer
to others,
making them a potential large-scale threat to real-life systems.

Since most of these attacks use neural network gradient information
to generate perturbations~\citep{goodfellow2014explaining},
the obvious defense is to improve the networks' classification robustness,
such as training classifiers with these adversarial images.
Such {\em adversarial training} significantly increases
the performance of CNNs on adversarial samples but falls short in three ways.
First,
it assumes the defender has prior knowledge of the attacks;
second, the defense is not certifiable;
third, building a fully robust model is still an unsolved question \citep{schott2018towards}.
In this paper, we look at a different defense strategy, namely adversarial sample detection.
Researchers have shown that many adversarial samples are detectable, and
detection methods normally hold no prior knowledge of attackers
\citep{meng2017magnet, shumailov2018taboo}.
We built on the existing
Taboo Trap detection scheme \citep{shumailov2018taboo},
whose focus is on finding overly excited neurons being driven out-of-bound
from a pre-defined range by adversarial perturbations.
We propose a mechanism, the Certifiable Taboo Trap (CTT), that combines the original Taboo Trap detection with numerical
bound propagation, making the detection bounds
on CNN activation values certifiable against certain input perturbation sizes.
For input perturbations at a particular
$l_{\infty}$ value, CTT can verify detection,
meaning that CTT guarantees the detected samples are adversarial inputs.

In this paper, we propose three versions of CTT: lite, loose and strict.
CTT-lite requires no additional fine-tuning on a pretrained model,
and can provide basic protection against weak adversaries.
CTT-loose retrains on a random
set of selected activations with propagated numerical interval bounds, and provides a loose guarantee that
all samples detected are adversarial.
Finally, CTT-strict fine-tunes with more strict numerical interval bounds
and thus is able to provide the same guarantee as CTT-loose on
attackers with small $l_{\infty}$ values; in addition, CTT-strict
can verify detection on a pre-defined range of $l_{\infty}$ values.

The contributions of this paper are:
\begin{itemize}
    \item We introduce a novel certifiable detection scheme for
    adversarial samples.
    \item We show CTT-lite,
    a new detection method that is fine-tuning free
    but relatively limited in its defense capability.
    We demonstrate how to optimise detection boundaries through
    fine-tuning and introduce CTT-loose and CTT-strict.
    Assuming the test and training data distributions are the same, both detection schemes
    ensure that all detected samples are adversarial,
    CTT-strict even guarantees detections on adversarial samples
    with particular range of pre-defined $l_{\infty}$ bounds.
    \item We show the detection results on all versions of CTT.
    For the first time, we empirically demonstrate
    how certifiable detection scheme (CTT-loose and CTT-strict) can have above $90\%$
    detection ratios on all attacks experimented on MNIST.
\end{itemize}


%% file: sections/related.tex
\section{Related Work}
\label{sec:rw}

The field of adversarial machine learning has seen a rapid co-evolution of attack and defense since researchers
discovered adversarial samples~\citep{Szegedy2013}.
The fast gradient sign method (FGSM) is an early
adversarial attack that generates perturbations using the signs of the network gradients, and is still a simple yet effective way of finding adversarial samples~\citep{goodfellow2014explaining}.
The FGSM attack can be extended in an iterative way to look
for smaller perturbations, giving the
Projected Gradient Descent (PGD) Method or the
Basic Iterative Method (BIM)~ \citep{kurakin2016adversarial, madry2017towards}.
The Carlini \& Wagner attack (CW) formulates
an optimization problem,
whose solution gives an adversarial sample~\citep{carlini2017towards}.
However, a strong adversarial image is time-consuming to generate since it requires
a large number of search iterations and binary search
steps.
Many of the attacks can change their optimization focus or be constrained on certain $l_p$ norms in an iterative run.
In our setup, we use the term $l_p$-bound attack to differentiate the
same attack bounded by various $l_p$ norms.

An interesting feature of adversarial samples is their transferability
\citep{Szegedy2013,zhao2018compress,goodfellow2014explaining}.
Adversarial samples that work well on a given neural network often
transfer to a different type of network trained on the same dataset.
This makes black-box adversarial attacks possible.
Another way of finding black-box attacks is the gradient estimation method, which uses
estimated gradient information instead of true gradients \citep{bhagoji2018black}.
Estimation involves building an output distribution based on information queried from the target model.

Many defenses against adversarial attacks have been proposed, most of them relying on improving
classification robustness.
Adversarial training adds adversarial samples
to the training set, so that the model becomes more robust at classification boundaries~\citep{goodfellow2014explaining, kurakin2016adversarial}.
Pang \etal use an ensemble of models to increase decision robustness~\cite{pang2019improving}, while Mustafa \etal use class-wise disentanglement to restrict feature maps
crossing the decision boundaries~\cite{mustafa2019adversarial}.
However, Schott \etal showed that even building robust classification
on the small MNIST data remains an unsolved question~\citep{schott2018towards}.
They also proposed the analysis and synthesis (ABS) method using
class-conditioned data and demonstrate better robustness on the MNIST classification
task.

Many researchers have tried to detect adversarial samples~\citep{meng2017magnet,lu2017safetynet,metzen2017detecting,shan2019gotta}.
Magnet claims that detection is possible
by inspecting the reconstruction error of a trained autoencoder~\citep{meng2017magnet}.
SafetyNet proposed SVM classifiers
to recognize adversaries through neural activation patterns~\citep{lu2017safetynet}.
However, both of these detection methods rely on auxiliary components, which have two main problems.
First, they impose a significant computational overhead.
Second, an adversary might obtain a copy of the defense and devise an adversarial sample to defeat it~\citep{carlini2017magnet, chen2019stateful}.

Another efficient detection scheme is the Taboo Trap \citep{shumailov2018taboo},
where a random subset of neurons are constrained in training and an alarm is set off when some threshold of them become overly excited.
This imposes no extra runtime computational costs, and the constrained subset of neurons can be randomly picked, giving what amounts to a key that can be different each time the network is trained.
This makes Black-box attacks more challenging as there can be multiple independently-keyed networks each of which is vulnerable to different adversarial samples~\citep{shumailov2019sitatapatra}.
Our work builds on the Taboo Trap, and answers the
question of how to make adversarial sample detections certifiable. It also estalishes
the optimal numerical range limit on neurons, and thus significantly
improves the detection performance of the Taboo Trap.

Our work can also be viewed as being related to
certifiable robustness where
the prediction of a data point $x$ is verifiably constant
with perturbations of a certain $l_p$ norm.
When queried with the input data $x$, $x$ will be perturbed by isotropic Gaussian noise and multiple inference runs are executed
on a base classifier $f$ \citep{cohen2019certified,lcuyer2018certified},
in this way, the returned classification provides the most probable prediction
made by $f$ with a Gaussian corrupted $x$.
Meanwhile, certification of adversarial samples can be achieved
using bound interval propagation, which is becoming established as a means of formal verification
of neural networks
\cite{gowal2018effectiveness,
    dvijotham2018training,
    wang2018efficient,
    wong2018scaling,
    gehr2018ai2,mirman2018differentiable,katz2017reluplex}.
Several prior works have studied efficient relaxation methods
for computing tight bounds on the neural network outputs \cite{wang2018efficient,wong2018scaling}.
Our Certifiable Taboo Trap uses bound-interval propagation,
but its focus is on certifying out-of-bound values in
a set of randomly sampled intermediate activations.
The interval bounding is a simple integral
bound so the computation overhead is minimised \cite{mirman2018differentiable}.




%% file: sections/method.tex
\section{Method}\label{sec:meth}

\begin{figure*}[!tbp]
  \centering
  \begin{subfigure}[b]{0.28\linewidth}
    \includegraphics[width=\linewidth]{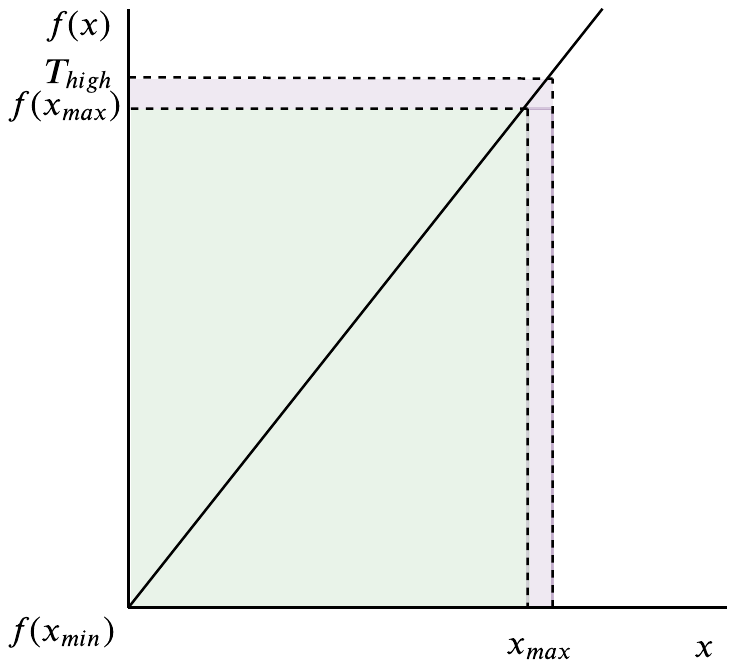}
    \caption{Original Taboo Trap.}
    \label{fig:taboo}
  \end{subfigure}
  \hfill
  \begin{subfigure}[b]{0.28\linewidth}
    \includegraphics[width=\linewidth]{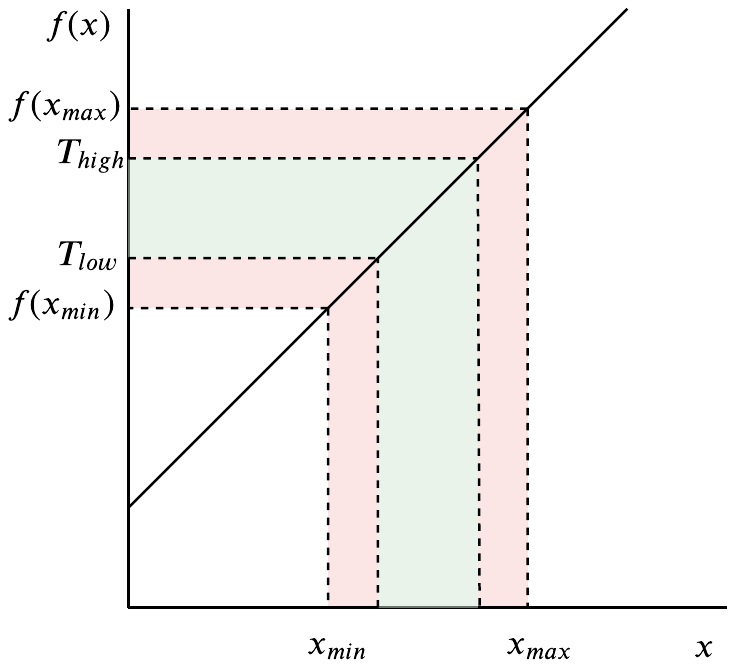}
    \caption{False positives (in red).}
    \label{fig:fp}
  \end{subfigure}
  \hfill
  \begin{subfigure}[b]{0.28\linewidth}
    \includegraphics[width=\linewidth]{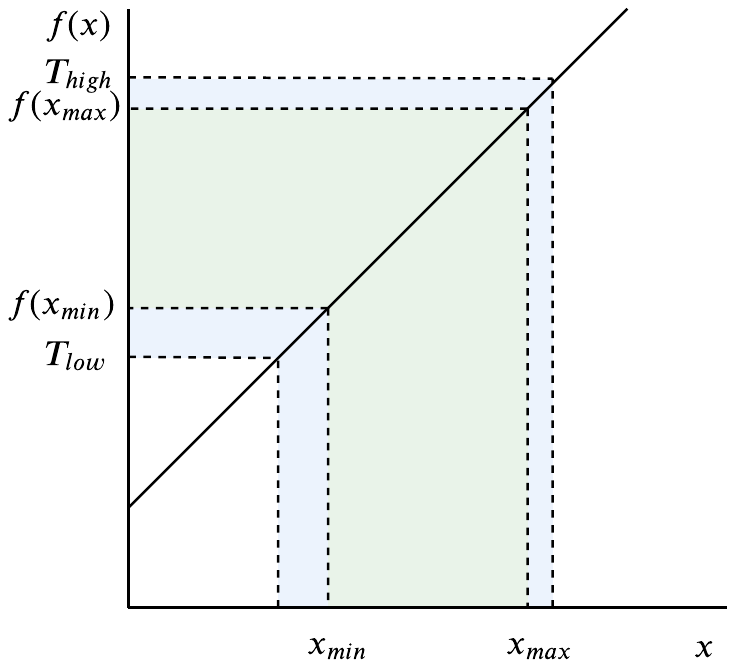}
    \caption{Undetectable range (in blue).}
    \label{fig:undetect}
  \end{subfigure}
  \caption{Taboo Trap visualisation.  If $x$ is from the original data distribution,
    a strict bound causes them to be detected as adversarial samples, the red
    area shows the false positive samples (middle figure).
    If $x$ is an adversarial sample, loose bounds cause detection fail on
    adversarial samples with small $l_{\infty}$ values (blue part in figure on the right).}
  \label{fig:vtt}
\end{figure*}




\subsection{Taboo Trap: A Practical View}

The method shown here extends the Taboo Trap originally
presented by~\citep{shumailov2018taboo, shumailov2019sitatapatra}.
First, we will explain the Taboo Trap method
and then demonstrate the extension made for producing a relaxed guarantee that a
certain $l_{\infty}$ bound attacker will always be detected.

The Taboo Trap is based on the idea that neural network activations
can be forced to form a distribution when trained with extra
regularisations.
Regularisations are based on activation values, and
bound a set of activations inside a
certain numerical range. No training set inputs trigger this chosen set
of activation values to be out of range.
So if one of these `taboo' activations is observed, it
signals that the current input may be adversarial.
As different instances of the model can be trained
with different taboo sets, the authors coined a term of a
\emph{transfer function}, which essentially served as a neural network key.
In the original Taboo Trap, Shumailov \etal made use of
the $n$th-max percentile activation bounds profiled from a trained network~\citep{shumailov2018taboo}.
They later used polynomial keys~\citep{shumailov2019sitatapatra}.
Yet, the detection rates reported were less than ideal:
the $n$th-max percentile function only detects weak attackers, while
polynomial-based detectors show good detection rates on transfer attacks but
perform worse under direct attack.

The Taboo Trap authors hypothesised that its performance is related to the
choice of transfer functions, yet could not explain
why some attackers could not be detected.
While their experiments show a practical ability to detect adversaries, there is little theoretical understanding of how and why it worked.

\subsection{Taboo Trap: A Theoretical View}
\label{sec:vtt:theo}
In this section, we provide a theoretical understanding of
how operating on the high dimensional activation space can detect adversarial samples.
Assume that we have a linear function $f(x) = ax + b$ for illustration simplicity.
The simple integral bound of the linear function
with input bounded between $x_{\tmin}$ and $x_{\tmax}$ is
bounded by $f(x_{\tmin})$ and $f(x_{\tmax})$.

\Cref{fig:taboo} presents how the original Taboo Trap
will instrument function $f$ with a $n$th max percentile transfer function.
$x_{\tmin}$ and $x_{\tmax}$ represent the minimum and maximum values
$x$ can take.
Since network inputs are bounded,
the intermediate layers should receive inputs
that are also bounded, regardless of non-linearities.
Being monotonic functions, $f(x_{\tmin})$ and
$f(x_{\tmax})$ present the minimum and maximum values that
the function $f$ can naturally assume.
If $T_{\high}$ represents the Taboo Trap
threshold; we have:

\begin{equation}
\begin{cases}
      f(x)\leq T_{\high} & \mathrm{Benign} \\
      f(x) > T_{\high} & \mathrm{Malicious} \\
\end{cases}
\end{equation}

We define an adversarial sample $\hat x = x+\epsilon$, with its
$l_{\infty}$ norm having the size of $\epsilon$.
With different detection thresholds ($T_{\high}$),
we can have natural samples becoming false positives or
adversarial samples becoming undetectable.
\Cref{fig:fp} shows the scenario when
$T_{\high} < f(x_{\tmax})$: there exists
a clean sample $x$ with an output
$f(x)$ being in between $T_{\high}$ and $f(x_{\tmax})$.
This triggers natural samples to be misclassified as adversarial (false positives).
\Cref{fig:undetect} presents the case that
$T_{\high} > f(x_{\tmax})$: adversarial samples $\hat x$ can
generate output $f(\hat x)$ smaller than $T_{\high}$ so that
it becomes undetectable by the Taboo Trap framework.
In summary:

\begin{equation}
\begin{cases}
	  T_{\high} > f(x_{\tmax}) & \text{Missed detection}\\
	  T_{\high} < f(x_{\tmax}) & \text{False positives}\\
  \end{cases}
\end{equation}

Consider $r = |f(x_{\tmax})-T_{\high}|$, it means
\begin{itemize}
	\item if $r$ equals to zero, the adversarial samples will always get detected.
	\item for a given $r$ it is easy to compute what type and how many of perturbations will go undetectable.
	\item as mentioned by Shumailov~\etal, there is a direct measurable trade-off between false positives, accuracy and detection rate.
\end{itemize}

Using the method defined above, it becomes apparent that
all monotonic transfer functions should theoretically work in Taboo Trap, and have
a trade-off between accuracy, false positive and detection rates.

Perturbations can also exist in the range between
$x_{\tmin}$ and $x_{\tmax}$. The original Taboo Trap paper
observed that better detector performance is achieved by setting a
small threshold value, yet training becomes hard. Our hypothesis is that
reducing the distance between $x_{\tmin}$ and $x_{\tmax}$ leads to a reduced
number of perturbations in the natural image range.

It is also worth noting that detection occurs on post-ReLU activation
values, and only the positive numerical range and the positive
numerical threshold ($T_{\high}$) are considered.
For simplicity, we use $T_l$ to represent a layer-wise
threshold scalar in later descriptions.

\subsection{Interval Bound Propagation}
\label{sec:ibp}

For simplicity, we consider a feed-forward CNN \( F \)
consisting of a sequence of convolution layers,
where the \( \ordinal{l} \) layer
computes output feature maps \(
\tx_{l} \in \realset^{C_l \times H_l \times W_l}
\).
\( \tx_{l} \) is a collection of feature maps
with \( C_l \) channels of \( H_l \times W_l \) images.

The first stage of CTT is to compute the activation bounds
from a pretrained network.
Given the pretrained weights and the numerical bounds of
inputs, CTT computes the numerical bounds for each
layer in the CNN.
Assuming the a set of lower and upper bound for layer $l$ is
$(B_l^{\low}, B_l^{\up})$,
where $B_l^{\low}$ is the lower bound and $B_l^{\up}$
is the upper bound
respectively, we have

\begin{equation}
  \begin{split}
    & B_{l+1}^{\low} =  \mathsf{Conv_b}(W_l, B_{l}^{\low})\\
    & B_{l+1}^{\up} = \mathsf{Conv_b}(W_l, B_{l}^{\up})
  \end{split}
\label{equ:bound}
\end{equation}

Notice $B_{0}^{\low}$ and $B_{0}^{\high}$ will be the boundaries
on the input, obtained from profiling on the natural input
data samples.
Both $B_l^{\low}$ and $B_l^{\up}$ have the same dimensions as $\tx_l$.
The interval bound propagation in \Cref{equ:bound}
can be seen as a series of abstract interpolations in a convolution
($\mathsf{Conv_b}$)
\citep{wong2018scaling,mirman2018differentiable}.
Given scalar bounds $m_l \leq m \leq m_h$ and $n_l \leq n \leq n_h$,
we define an operation $(m_l, m_h) \cap (n_l, n_h)$
produces a tighter bound $p_l = \mathrm{max}(m_l, n_l)$
and $p_h = \mathrm{min}(m_h, n_h)$:

\begin{equation}
  (m_l, m_h) \cap (n_l, n_h) = (\tmax(m_l, m_l), \tmin(n_h, n_h))
\end{equation}

\begin{figure*}[!h]
  \centering
  \hspace*{\fill}%
  \begin{subfigure}[h]{0.3\textwidth}
    \centering
    \includegraphics[width=\linewidth]{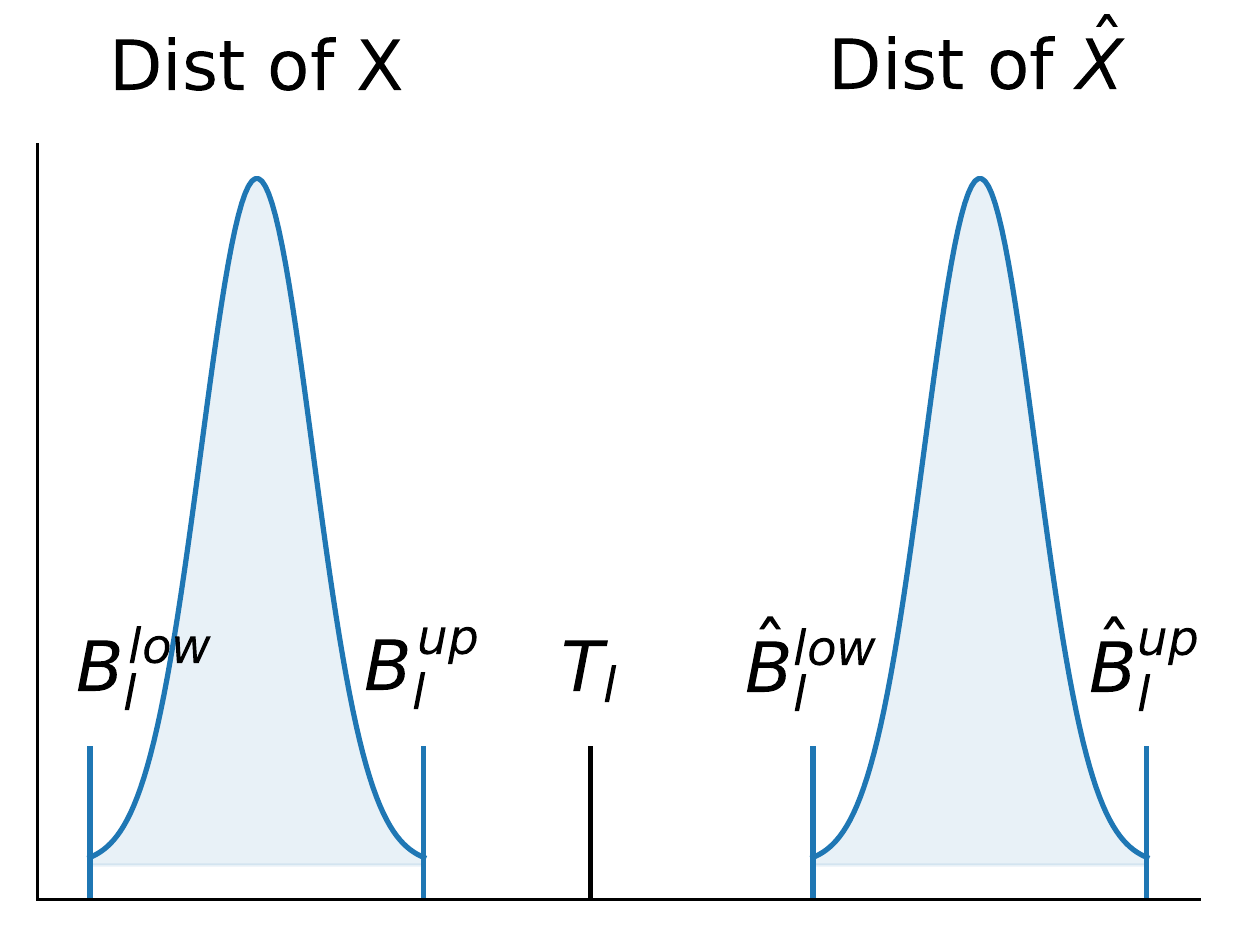}
    \caption{Optimal placement of $T_l$.}
    \label{fig:ideal}
  \end{subfigure}
  \hfill
  \begin{subfigure}[h]{0.3\textwidth}
    \centering
    \includegraphics[width=\linewidth]{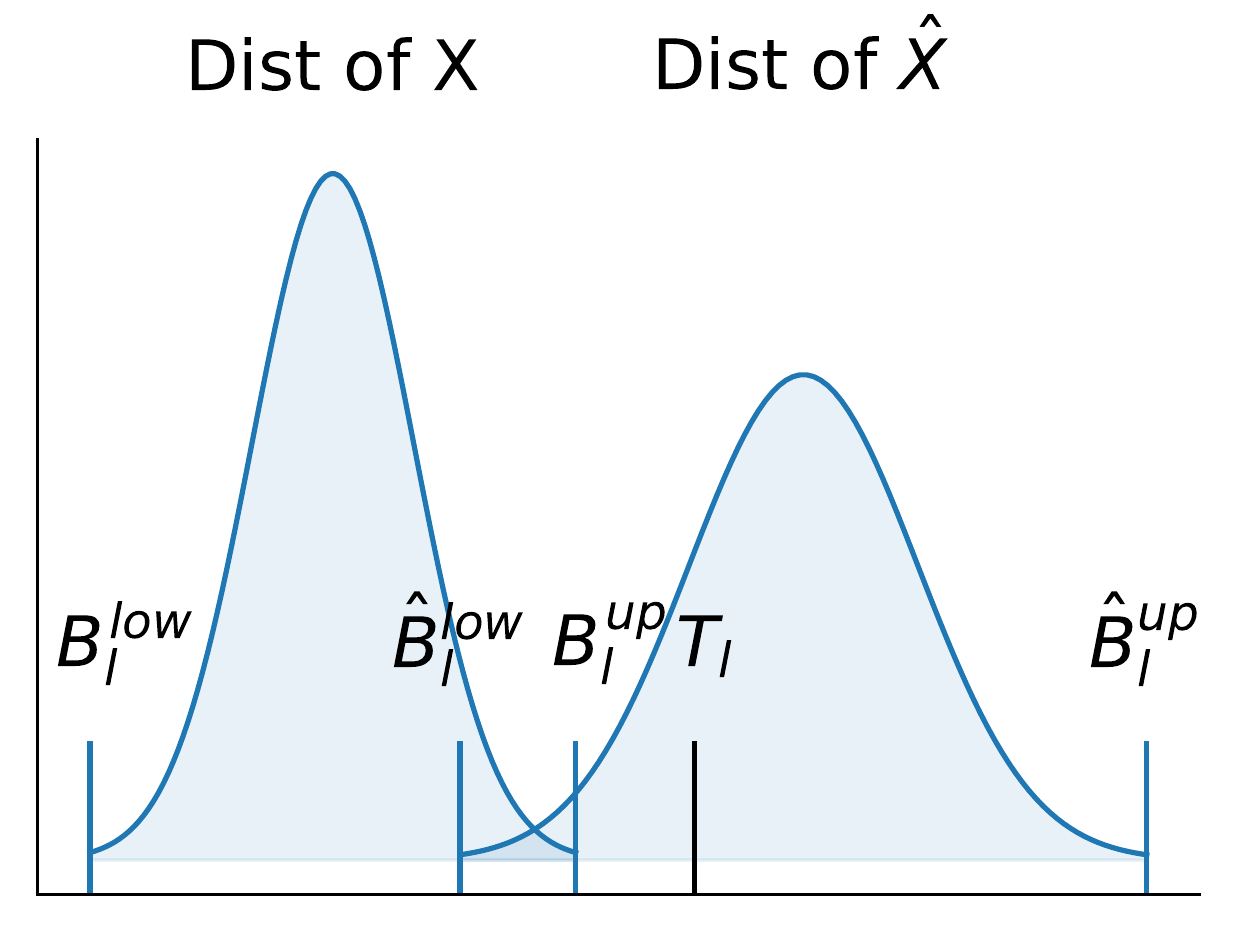}
    \caption{Suboptimal placement of $T_l$.}
    \label{fig:real}
  \end{subfigure}
  \hspace*{\fill}%
  \caption{Placements of the detection threshold $T_l$ with different
  boundaries from both the natural activations ($X$) and the adversarial activations ($\hat X$).
  This indicates that both the numerical value of $T_l$ and the two distributions
  should be optimised using fine-tuning.}
  \label{fig:placement}
\end{figure*}

This is equivalent to abstract interpolation in
the interval domain (the box domain)~\cite{mirman2018differentiable}.
Notice the bound propagation can also be performed for
adversarial inputs, where the upper bound of the input becomes
$\hat B^{\up}_{0} = B^{\up}_{0} + \epsilon$ and $\epsilon$
is the size of the $l_{\infty}$ norm.
We then define $(\hat B_l^{\low}, \hat B_l^{\up})$ to be a pair of
upper bound and lower bound for layer $l$ for adversarial inputs with an
$l_{\infty}$ budget of $\epsilon$.

Considering the case in the middle layer $l$, we obtain a particular activation
value $x$ and its values across all input data distributions can be seen
as a set $X$. Meanwhile, its values for all adversarial samples with
an adversarial perturbation can be viewed as a set $\hat X$.
For convenience, we call $X$ the natural set and $\hat X$ the adversarial set.
\Cref{fig:placement} shows the placements of the detection threshold
$T_l$.
In the ideal case, if the distributions of the natural set ($X$) and
the adversarial set ($\hat X$) are disjoint, the optimal
placement of $T_l$ is that $B^{\up}_{l} <= T_l <= \hat B^{\low}_{l}$.
However, in practice, the natural set and the adversarial
set might overlap (\Cref{fig:real}), meaning that there is only a sub-optimal placement
option $B^{\up}_{l} \leq T_l \leq \hat B^{\up}_{l}$.
For these two threshold placements, we conclude:
\begin{itemize}
  \item Optimal placement of $T_l$ (
  $ B^{\up}_{l} \leq T_l \leq \hat B^{\low}_{l}$)
  ensures that all adversarial samples with
  $l_{\infty}$ norm at the size of $\epsilon$ are detectable (\Cref{fig:ideal}).
  \item Both optimal and suboptimal placements
  ($ B^{\up}_{l} \leq T_l \leq \hat B^{\up}_{l}$)
  of $T_l$
  ensure that all detected samples are adversarial regardless of
  the perturbation size (\Cref{fig:real}).
\end{itemize}
The above claims are true
if and only if the following assumption holds:
\emph{The test data distribution falls inside the training data distribution}.
In other words, the test data falls in the range of the
maximum and minimum bounds profiled from the training dataset.
The rest of this section discusses methodologies we used
to ensure that placements of $T_l$ are near-optimal.
In \Cref{sec:vtt:free}, we show a training-free method of finding
the position of $T_l$.
In \Cref{sec:vtt:loss}, several losses are discussed
that help to force the placement of $T_l$ towards optimal and suboptimal.

\begin{algorithm}[!h]
\caption{Certifiable Taboo Trap finetuning process}
\label{alg:training}
\begin{algorithmic}
\STATE {\bfseries Inputs:} $\alpha$, $\beta$, $\theta$, $f$, $x$, $y$, $\epsilon$, $E$
\STATE $\mathbf{m}^d = \mathrm{RandomMaskGen}(\beta)$
\FOR{$e=0$ {\bfseries to} $E - 1$}
  \STATE $L = \mathrm{CrossEntropy}(y, f(x))$
  \STATE $B = \emptyset, \hat B = \emptyset$
  \FOR{\texttt{$l \in \mathrm{Layers}(f)$}}
    \STATE $(B^{\up}_{l}, B^{\low}_{l}) = \mathrm{BoundPropagate}(l, x, f)$
    \STATE $(\hat B^{\up}_{l}, \hat B^{\low}_{l}) = \mathrm{BoundPropagate}(l, x\pm\epsilon, f)$
    \STATE $B = B \cup (B^{\up}_{l}, B^{\low}_{l})$
    \STATE $\hat B = \hat B \cup (\hat B^{\up}_{l}, \hat B^{\low}_{l})$
  \ENDFOR
  \STATE{$L_{D}, L_{V} = \mathrm{ComputeRegLoss}(B, \hat B, f(x), \mathbf{m}^d)$}
  \STATE{$\alpha =\mathrm{Anneal}(\alpha, e)$}
  \STATE $\mathsf{Opt}_{\theta} (L + \alpha(L_D + L_V))$
\ENDFOR
\end{algorithmic}
\end{algorithm}

\subsection{Taboo Trap for Free}
\label{sec:vtt:free}
One major bottleneck of defending adversarial samples is
the training overhead.
Classic methods like adversarial training increase model robustness by training
with additional adversarial data points and thus significantly
increase the training time.
CTT can be deployed without any additional
fine-tuning, and we name this detection mode CTT-lite.

We previously introduced the concept of a detection threshold value.
Recall the definition of a particular layer's output activations
$\tx_{l}$, CTT uses a randomised binary mask $\tm^d_{l}$ that is the same
size of $\tx_{l}$ to decided on which activation values to restrict on.
Unlike \citep{shumailov2019sitatapatra} who used different transfer
functions as keys, in this work we represent different keys as different subsets of neurons that are instrumented with CTT.
We find that such construction has all of the benefits described by \citep{shumailov2019sitatapatra} originally.
Practically, CTT only detects on $\tx_{l} \cdot \tm^d_{l}$, where $\cdot$ is
a Hadamard product (element-wise multiplication) between matrices.

CTT-lite simply
places $T_l$ at the upper boundary of the natural set so that $T_l=B^{\up}_{l}$.
In the original Taboo Trap setup, as in \Cref{sec:vtt:theo},
this effectively means $r=|f(x_{\tmax})-T| = 0$.
So the only additional computation is to perform the interval
bound propagation for deducing the value of $T_l$ in each layer, and no additional
training is required. Note that as the bounds are computed for the training dataset it will have false positives for the evaluation dataset.

\begin{figure*}[!h]
  \begin{subfigure}[h]{0.32\textwidth}
    \includegraphics[width=\linewidth]{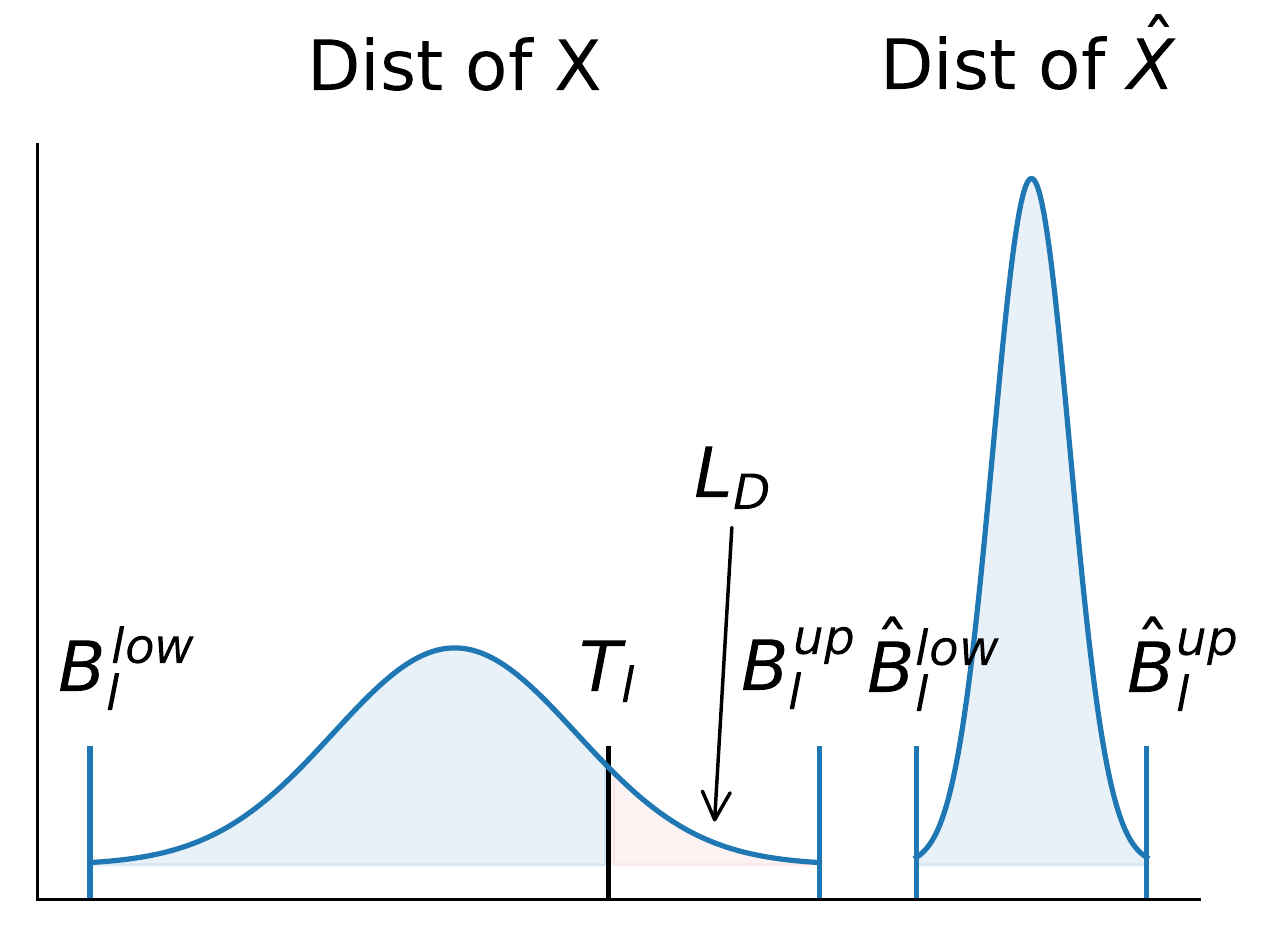}
    \caption{Detection loss $L_{D}$.}
    \label{fig:detect}
  \end{subfigure}
  \begin{subfigure}[h]{0.32\textwidth}
    \includegraphics[width=\linewidth]{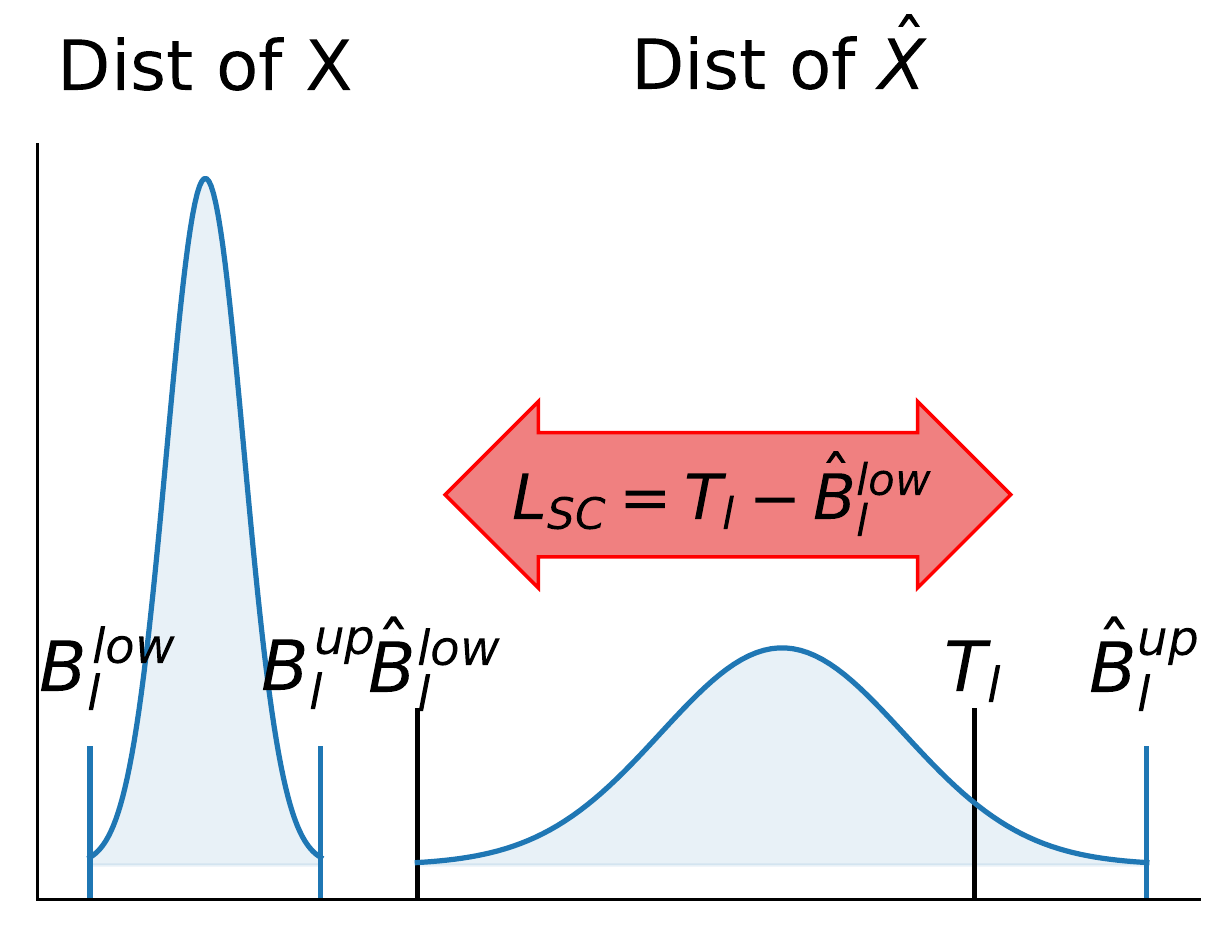}
    \caption{Strict Certification loss $L_{SC}$.}
    \label{fig:strict}
  \end{subfigure}
  \begin{subfigure}[h]{0.32\textwidth}
    \includegraphics[width=\linewidth]{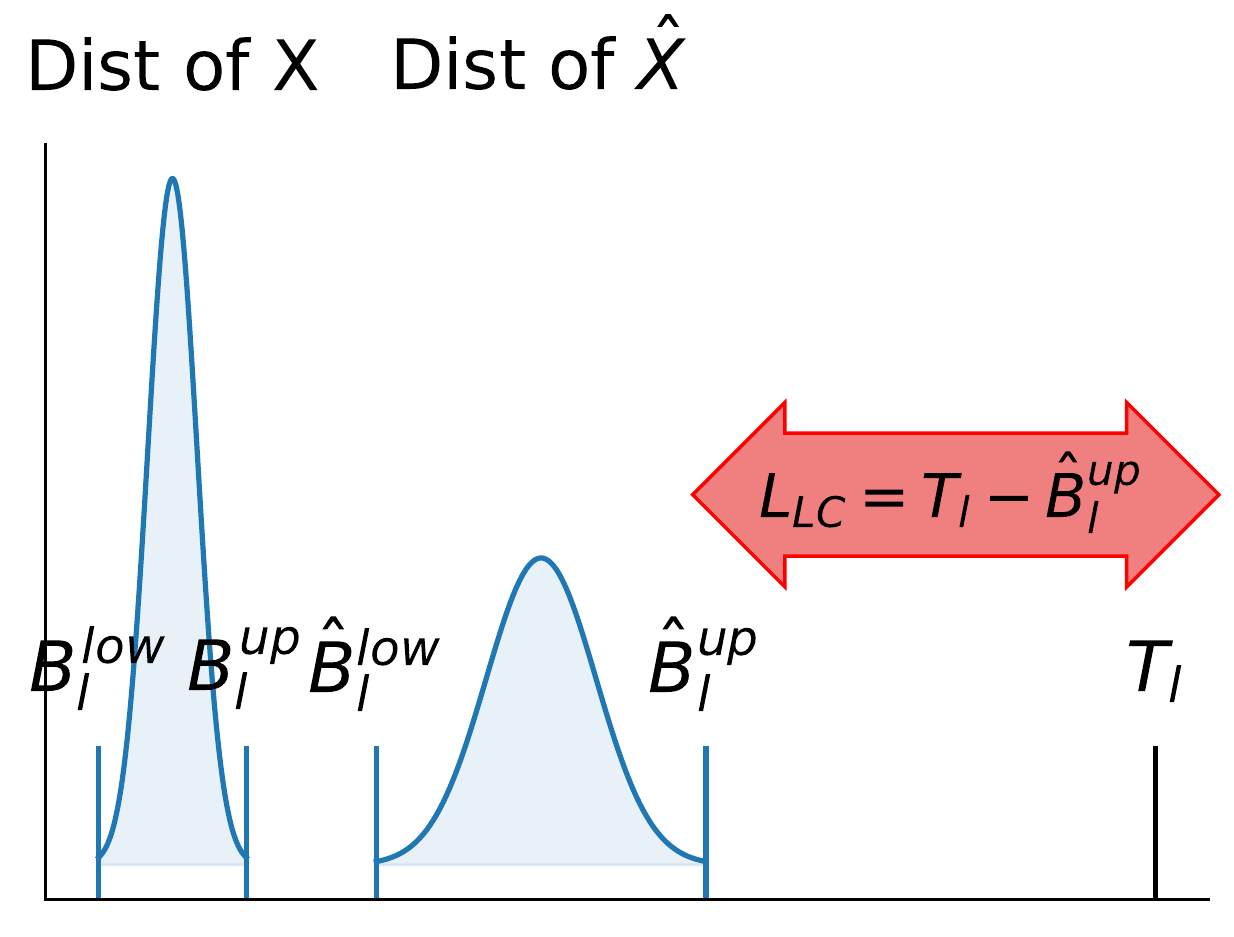}
    \caption{Loose Certification loss $L_{LC}$.}
    \label{fig:loose}
  \end{subfigure}
  \caption{
  An illustration of CTT regularisation losses.
  The detection loss ($L_D$) ensures no natural samples are detected.
  Strict certification loss encourages the placement of $T_l$ to be optimal,
  while loose certification loss helps $T_l$ to achieve the suboptimal placement.}
  \label{fig:vtt:loss}
\end{figure*}


\subsection{Fine-tuning with CTT Losses}

\label{sec:vtt:loss}
Fine-tuning networks further with CTT losses can introduce a better separation
between the natural and the adversarial sets.
Unlike adversarial training, CTT fine-tuning operates on the original data; we do not generate any adversarial inputs to train with the model, so the training overheads are lower for CTT.
We present three losses related to interval bounds that are considered
as regularisations in our CTT detection.
The three losses are presented in \Cref{fig:vtt:loss}, and they are:
1) Detection loss $L_D$, 2) Strict certification loss $L_{SC}$,
3) Loose certification loss $L_{LC}$.

Consider a masking function $\mathbf{m}_l=M(\mathbf{x}_l, T_l)$,
the output $\mathbf{m}$ is a binary mask of which its elementwise entry
is $1$ if its corresponding elementwise entry in $\mathbf{x}$ is bigger
than a scalar $T_l$, and otherwise is $0$.
The detection loss $L_D$ is a sum of all activation values picked by
the taboo selection mask $\tm^d_l$ that are greater than
the detection threshold $T_l$.
The verification losses are simply the distance between the detection
threshold and the bound when the threshold is bigger than the bound.
Considering a network with $N$ layers, we have:

\begin{equation}
  L_D = \sum^{N-1}_{l=0} \mathsf{sum}(\tx_l \cdot \tm^d_l \cdot M(\mathbf{x}_l, T_l))
\end{equation}
\begin{equation}
  L_{SC} = \sum^{N-1}_{l=0} \mathsf{sum}(\tm^d_l \cdot M(\hat B^{\low}_{l}, T_l) \cdot (T_l - \hat B^{\low}_{l}))
  \label{equ:sv}
\end{equation}
\begin{equation}
  L_{LC} = \sum^{N-1}_{l=0} \mathsf{sum}(\tm^d_l \cdot M(\hat B^{\up}_{l}, T_l) \cdot (T_l - \hat B^{\up}_{l}))
  \label{equ:lv}
\end{equation}

The function $\mathsf{sum}$ produces the sum of all entries
of a high dimensional tensor that is the result of convolutions (activations).
Recall we previously defined the optimal and suboptimal
placements of $T_l$ in \Cref{sec:ibp}, the minimization
of different combination of CTT regularisation losses provide:
\begin{itemize}
  \item If $L_D = 0$ and $L_{SC} = 0$,
  we are achieving the optimal placement of $T_l$.
  All adversarial inputs with its $l_{\infty}$ norm equals to $\epsilon$
  are detectable, and
  all detected samples are adversarial samples regardless of the perturbation size.
  Given that test data falls into the train data distribution.
  \item If $L_D = 0$ and $L_{LC} = 0$,
  we are achieving the suboptimal placement of $T_l$,
  all detected samples are adversarial samples regardless of the perturbation size.
  Given that test data falls into the train data distribution.
\end{itemize}
\input{tables/vtt_mnist.tex}

We present the detailed fine-tuning algorithm in \Cref{alg:training}.
The finetuning function takes a hyperparameter $\alpha$, this controls
how strong the regularisation is in the optimization procedure ($\mathrm{Opt}$).
In practice, it is necessary to anneal ($\mathrm{Anneal}$) the value of
$\alpha$ with respect to the number of epoch $e$.
The other hyperparameter $\beta$ is a probability between $0$ to $1$ that
is later used to produce a set of masks $\mathbf{m}^d$ for each layer's
activations.
In the meantime, the fine-tune function considers a neural network $f$
with trained parameters $\theta$; $x$ and $y$ are the training data samples
and their labels respectively.
In addition, we need a pre-defined perturbation size $\epsilon$ for
adversarial bound construction and $E$ represents the maximum number
of epochs we would like to fine-tune for.
Function $\mathrm{CrossEntropy}$ essentially computes the classification loss $L$
based on the input training data.

Consider a neural network $f$ parameterised by $\theta$.
For each layer in the neural network $f$, we perform the bound propagation
($\mathrm{BoundPropagate}$)
as described in \Cref{sec:ibp}.
The bounds for both the adversarial set of inputs and the natural set of inputs
of each layer are accumulated for computing the regularisation loss using
the function $\mathrm{ComputeRegLoss}$.
Note that the adversarial set represents the set of inputs
with a particular $l_{\infty}$ norm, so there is no actual generation of
adversarial samples.
The function $\mathrm{ComputeRegLoss}$ produces two losses
$L_D$ and $L_C$; the value of $L_C$ can be calculated to be equal
to whether $L_{SC}$ or $L_{LC}$ (\Cref{equ:sv} and \Cref{equ:lv}) depending
on whether we use CTT-strict or CTT-loose.
Since \Cref{alg:training} is only a high level overview, we did not
distinguish between $L_{SC}$ and $L_{LC}$, but call them in general
$L_C$ in \Cref{alg:training}.
It is worth to note that $L_{SC}$ is a stronger regularisation than
$L_{LC}$, so adding both regularisations is theoretically equivalent to
adding only $L_{SC}$. The pre-defined parameter $\epsilon$
determines a trade-off between accuracy,
detection ratios and adversarial accuracy.
In practice, we determine the value of $\epsilon$ using
a grid search spanning values from $10^{-5}$ to $10^{-1}$,
and determine its value based on the optimal performance in accuracy and detection ratio
under a simple FGSM attack with fixed $l_0$.
We explain this trade-off in details in our supplimentary material.

%% file: tables/vtt_mnist.tex
\begin{table*}[!h]
\centering
\caption{
    A comparison between CTT-lite, CTT-loose, CTT-strict,
    AdvTrain \cite{kurakin2016adversarial},
    Ensemble \cite{pang2019improving} and
    PCL \cite{mustafa2019adversarial} on the MNIST dataset.
    Acc means accuracy and Det means detection rate on adversarial
    samples.
    }

\begin{adjustbox}{scale=0.8,center}
\begin{tabular}{@{}llllll|lll|lll|lll@{}}
\toprule
&
& Baseline
& AdvTrain
& Ensemble
& PCL
& \multicolumn{3}{c}{CTT-lite}
& \multicolumn{3}{c}{CTT-loose}
& \multicolumn{3}{c}{CTT-strict}
\\
Attack
& Param
& Acc
& Acc
& Acc
& Acc
& Acc     &Det     &$l_2$
& Acc     &Det     &$l_2$
& Acc     &Det     &$l_2$
\\ \midrule
No Attack
&
& 99.1
& 99.5
& 99.5
& 99.3
& 99.1    & 1.9    &-
& 98.5    & 1.6    &-
& 98.9    & 1.1    &-
\\ \midrule
\multirow{2}{*}{FGSM}
& $\epsilon=0.1$
& 70.9
& 73.0
& 96.3
& 96.5
& 70.9    & 1.4    & 2.08
& 25.0    & 100.0  & 1.98
& 61.1    & 100.0  & 1.99
\\
& $\epsilon=0.2$
& 21.9
& 52.7
& 52.8
& 77.9
& 21.9 & 1.0 & 4.14
& 15.0    & 100.0  & 3.89
& 32.7    & 100.0  & 3.90
\\
\midrule
\multirow{2}{*}{BIM}
& $\epsilon=0.1$
& 44.2
& 62.0
& 88.5
& 92.1
& 44.2       & 1.0  & 1.13
& 0.0       & 100.0  & 0.38
& 0.15       & 100.0  & 0.75
\\
& $\epsilon=0.15$
& 4.2
& 18.7
& 73.6
& 77.3
& 4.2       & 0.8  & 1.48
& 0.0      & 100.0  & 0.50
& 2.0       & 100.0  & 0.97
\\
\midrule

\multirow{2}{*}{PGD}
& $\epsilon=0.1$
& 51.0
& 62.7
& 82.8
& 93.9
& 51.0       & 1.2  & 1.50
& 1.0     & 100.0  & 1.24
& 13.4       & 100.0  & 1.35
\\
& $\epsilon=0.2$
& 0.0
& 31.9
& 41.0
& 80.2
& 0.0       & 1.1  & 2.73
& 0.0     & 100.0  & 2.43
& 0.9       & 100.0  & 2.53
\\
\midrule
\multirow{2}{*}{C\&W}
& $c=0.1$
& 99.6
& 71.1
& 97.3
& 97.6
& 99.6       & 25.0 & 0.05
& 34.0      & 90.5  & 0.06
& 79.6       & 91.2  & 0.05

\\
& $c=1.0$
& 99.6
& 39.2
& 78.1
& 91.2
& 99.6       & 1.1  &0.05
& 34.3      & 93.3  & 0.07
& 79.7       & 96.1  & 0.06
\\
\bottomrule
\end{tabular}
\end{adjustbox}
\label{tab:mnist}
\end{table*}

%% file: sections/evaluation.tex
\section{Evaluation}\label{sec:eval}
\input{tables/vtt_cifar10.tex}

\subsection{Networks, Datasets and Attacks}
\input{tables/mnist.tex}
\label{sec:eval:setup}
We evaluate the proposed Certifiable Taboo Trap (CTT) on two different image datasets,
MNIST \citep{lecun2010mnist} and
CIFAR10 \citep{krizhevsky2014cifar}.
The MNIST dataset consists of images of hand-written digts and the number of output
classes is 10.
The CIFAR10 dataset is a task of classifying 60000 images into 10 classes.
We use the LeNet5 \citep{lecun2015lenet} architecture for MNIST,
and evaluate an efficient CNN architecture (MCifarNet) from Mayo \citep{zhao2018mayo}
that achieved a high classification accuracy using only 1.3M parameters.

We consider gradient-based FGSM~\citep{goodfellow2014explaining},
FGM~\citep{goodfellow2014explaining},
BIM~\citep{kurakin2016adversarial},
PGD~\citep{kurakin2016adversarial}
and C\&W~\citep{carlini2017towards}
attacks with various attack parameters.
These attacks can be seen as a collection of $l_{\infty}$ and $l_{2}$ based attacks.
In addition, we provide results in both White-box and Black-box settings.
For Black-box attacks,
we use gradient estimation with the coordinate-wise finite-difference method, similar to Schott~\etal~\citep{schott2018towards}.
The attack implementations are from Fooblox \citep{rauber2017foolbox}.

\subsection{Attackers with Various Capabilities and Various Norm Bounds}
Attacks can be evaluated very differently, and
we offer two sets of evaluations for a thorough comparison
with existing defense methods. In the first set,
we run attacks with fixed parameters and a fixed number of iterations.
In the second se, we enable early stopping for iterative
attacks so that perturbation sizes are fixed.
In addition, we also provide a set of evaluation
with Black-box attacks using gradient estimation.
We used $\epsilon = 3\times10^{-3}$ for MNIST networks, and $\epsilon = 10^{-4}$ for CIFAR10 networks.
These values were determined from a grid search;
there is a detailed discussion of the grid search and
an evaluation of using different $\epsilon$ in the supplementary material.
In addition, we show the detailed hyperparameter
configurations of $\alpha$, $\beta$, $E$ and $\epsilon$ (\Cref{alg:training})
in the supplementary material.

In \Cref{tab:mnist} and \Cref{tab:cifar10},
we present comparisons between CTT and various robust adversarial training schemes,
including AdvTrain~\cite{kurakin2016adversarial},
Ensemble~\cite{pang2019improving} and PCL~\cite{mustafa2019adversarial}.
In this setup, we run attacks with fixed parameters and measure
the accuracy, detection ratios and $l_2$ norms of the adversarial samples.
BIM and PGD iterated for 10 times with a step size of $\epsilon/10$;
CW has an iteration step of $1000$, an learning rate of $0.01$,
a confidence value of $0.1$ and a binary search step of $1$.
Notice we present the baseline accuracy for the networks on which we evaluate.
The baseline accuracy will be the same as CTT-lite, since it
involves no re-training of the model.
CTT-lite provides limited protections against
adversarial attacks.
CTT-loose and CTT-strict, however, show above $90\%$ detection ratios
across all examined attacks in \Cref{tab:mnist}.
In addition, both detection schemes provide
a degree of certifiability on the detected adversarial samples.
The detection ratios when no attacks are applied are the false
positives.
There exists a trade-off between the false positive rates
and the detection ratios.
As presented in \Cref{tab:cifar10},
the two versions of CTT-loose have different false positive rates,
and offer different detection capabilities.
We presents a full analysis of this trade-off in the supplementary material.
\Cref{tab:cifar10} shows our detection scheme outperform
robust networks on FGSM, however, provides relatively worse performance
when $l_2$ norms are low.
First, our detection offers certifiability which is not seen in any of the
work compared.
Second, the work compared does not report the $l_2$ norm,
attacks with different random starts may cause a difference in
$l_2$ norms and also the attacking quality.

To further evaluate the CTT system,
we conduct a comparison to \citeauthor{madry2017towards},
Sitatapatra \cite{shumailov2019sitatapatra}
ABS and Binary ABS \cite{schott2018towards}
under both White-
and Black-Box attacks on the MNIST dataset.
This time, we provide a noise budget to each attack.
The Black-Box attacks are constructed using gradient estimation,
we see almost all CTT-loose and CTT-strict results
show above $90\%$ detection ratios on adversarial samples while
keeping the false positives low.
Our detection results outperform all other competitors
focusing solely on improving model robustness.
An important observation is that our detection method reduces the
adversarial accuracy of the neural network model.
This phenomenon is apparent in \Cref{tab:mnist}, as all robustness-based
defenses have higher accuracy in comparison to CTT
on adversarial images.
Intuitively, CTT enforces
the natural and the adversarial sets to be separated by
the detection thresholds.
The CTT models
are thus more sensitive to adversarial samples because of this enforcement --- we observe that
selected neurons get suppressed for natural and get
non-zero values for adversarial inputs.
Furthermore, our detection is certifiable based on an
assumption that the test and train data follows the same
distribution.
In theory, if this assumption is true, CTT-strict will show
$100\%$ detection on all attacks
that are above a certain given $l_{\infty}$;
which is exactly the case shown in \Cref{tab:mnist} with
$l_{\infty}$-based attacks.
For $l_2$-based attacks, it is hard to
ensure every pixel is under the given certifiable limit,
however, our method practically capture many adversaries
with high detection rates.

\subsection{Runtime Overheads and Security Protocols}
The proposed CTT system has low running overheads in comparison
to other detection systems (SafetyNet \cite{lu2017safetynet}
and MagNet \cite{meng2017magnet}),
we present the run-time comparison in appendix.
It is similar to Sitatapatra~\cite{shumailov2019sitatapatra},
which is another derivative of Taboo Trap;
CTT supports the concept of embedding keys in each neural network
to diversify models under adversarial attack.
The key is embedded via the mask and can support complex
security protocols; a detailed analysis of key attribution and runtime overheads can
be found in~\citeauthor{shumailov2019sitatapatra} and
these advantages are equally
applicable to CTT.

%% file: tables/vtt_cifar10.tex
\begin{table*}[!h]
\centering
\caption{
    A comparison between CTT-loose, CTT-strict,
    AdvTrain \cite{kurakin2016adversarial},
    Ensemble \cite{pang2019improving} and
    PCL \cite{mustafa2019adversarial} on the Cifar10 dataset.
    Acc means accuracy and Det means detection rate on adversarial
    samples.
    }
\begin{adjustbox}{scale=0.8,center}
\begin{tabular}{@{}llllll|lll|lll|llll@{}}
\toprule
&
& Baseline
& AdvTrain
& Ensemble
& PCL
& \multicolumn{6}{c}{CTT-loose}
& \multicolumn{3}{c}{CTT-strict}
\\
Attack
& Param
& Acc
& Acc
& Acc
& Acc
& Acc     &Det     &$l_2$
& Acc     &Det     &$l_2$
& Acc     &Det     &$l_2$
\\ \midrule
No Attack
&
& 89.1
& 84.5
& 90.6
& 91.9
& 86.2    & 3.4    &-
& 86.3    & 6.4    &-
& 86.1    & 3.0    &-
\\ \midrule
\multirow{2}{*}{FGSM}
& $\epsilon=0.02$
& 33.6
& 44.3
& 61.7
& 78.5
& 18.6    & 95.7   & 1.07
& 16.8    & 98.5   & 1.08
& 16.1    & 96.4   & 1.06
\\
& $\epsilon=0.04$
& 22.4
& 31.0
& 46.2
& 69.9
& 7.6    & 93.6   & 2.00
& 7.2    & 94.2   & 2.01
& 6.0    & 93.1   & 2.06
\\
\midrule
\multirow{2}{*}{BIM}
& $\epsilon=0.01$
& 13.5
& 22.6
& 46.6
& 74.5
& 0.5 & 9.0    & 0.15
& 0.0 & 14.1   & 0.16
& 1.1 & 10.9   & 0.16
\\
& $\epsilon=0.02$
& 1.5
& 7.8
& 31.0
& 57.3
& 0.0     & 14.2   & 0.21
& 0.0     & 25.9   & 0.20
& 0.0     & 17.2   & 0.21
\\
\midrule

\multirow{2}{*}{PGD}
& $\epsilon=0.01$
& 24.0
& 24.3
& 48.4
& 75.7
& 0.1     & 10.4   & 0.34
& 2.9     & 24.3   & 0.34
& 2.0     & 16.6   & 0.34
\\
& $\epsilon=0.02$
& 2.9
& 7.8
& 30.4
& 48.5
& 0.0     & 40.8   & 0.65
& 0.0     & 70.3   & 0.65
& 0.0     & 49.9   & 0.65
\\
\midrule
\multirow{2}{*}{C\&W}
& $c=0.01$
& 13.5
& 40.9
& 54.9
& 65.7
& 0.2     & 12.9       &0.09
& 0.1     & 23.4       &0.10
& 0.5     & 14.5       &0.09
\\
& $c=0.1$
& 13.3
& 25.4
& 25.6
& 60.5
& 0.3       & 13.3       & 0.09
& 0.1       & 25.9       & 0.10
& 0.5       & 13.7       & 0.09

\\
\bottomrule
\end{tabular}
\end{adjustbox}
\label{tab:cifar10}
\end{table*}

%% file: tables/mnist.tex
\begin{table*}[!h]
\centering
\caption{
    A comparison between CTT-lite, CTT-loose, CTT-strict,
    Madry \etal \cite{madry2017towards},
    Sitatapatra \cite{shumailov2019sitatapatra}, ABS and
    Binary ABS \cite{schott2018towards} on the MNIST dataset.
    For detection based defense, we show results in the form of
    $a (d)$, where $a$ is accuracy and $d$ is detection rate.
    GE represents gradient estimation.
}

\begin{adjustbox}{scale=0.8,center}
\begin{tabular}{@{}llllll|ll@{}}
\toprule
                & CNN           & Madry \etal       & Binary ABS    & ABS            & Sitatapatra     & CTT-loose  & CTT-strict    \\
\midrule
No Attack       & $99.1\%$      & $98.8\%$          & $99.0\%$      & $99.0\%$       & $99.2\% \ (2\%)$  & $99.1\% \ (0.5\%)$ & $98.8\% \ (1.3\%)$ \\
\midrule
$l_2$-metric ($\epsilon = 1.5$)&&&&&&\\
FGM             & $48\%$      & $96\%$            & -             & -                & $2\%  \ (3\%)$  & $ 4\%  \ (99\%)$ & $21\% (100\%)$  \\
FGM w/ GE       & $42\%$        & $88\%$            & $68\%$        & $89\%$         & $4\%  \ (7\% )$   & $0\%  \ (100\%)$ & $25\% (100\%)$\\
Deepfool        & $18\%$        & $91\%$            & -             & -              & $12\% \ ( 1\%)$    &$ 0\% \ ( 100\%)$& $ 77\% \ (95.6\%)$ \\
Deepfool w/ GE  & $30\%$        & $90\%$            & $41\%$        & $83\%$         & $6\%  \ (2\% )$   & $0\%  \ (100\% )$ & $76.5\%  \ (94.4\% )$\\
L2 BIM             & $13\%$        & $88\%$            & -             & -           & $0\%  \ (0\% )$   & $0\%  \ (100\% )$ & $0\% \ (100\% )$\\
L2 BIM w/ GE       & $37\%$        & $88\%$            & $63\%$        & $87\%$      & $0\%  \ (3\% )$   & $0\%  \ (100\% )$ & $0\% \ (100\% )$\\
\midrule
$l_{\infty}$-metric ($\epsilon = 0.3$)&&&&&&\\
FGSM            & $4\%$        & $93\%$            & -             & -              & $2\% (3\%$)&  $1\%  \ (99\% )$ & $2\% (100\%)$ \\
FGSM w/ GE      & $21\%$       & $89\%$            & $85\%$        & $34\%$         & $0\% (2\%$) & $ 0\% \ ( 100\%)$ & $ 4\% \ ( 100\%)$  \\
BIM             & $0\%$        & $90\%$            & -             & -              & $0\% (1\%$) & $ 0\% \ ( 100\%)$ & $ 0\% \ ( 100\%)$  \\
BIM w/ GE       & $37\%$       & $89\%$            & $86\%$        & $13\%$         & $0\% (1\%$) & $ 0\% \ ( 100\%)$ & $0\% \ (100\% )$\\
\bottomrule
\end{tabular}
\end{adjustbox}

\label{tab:reg-effect}
\end{table*}

%% file: sections/conclusion.tex
\section{Conclusion}\label{sec:conc}

In this paper, we presented the Certifiable Taboo Trap CTT),
a new way to defend neural networks against adversarial samples by detecting them.
We discussed three different modes which provide different
detection capabilities and levels of certifiability
at different training costs.
All variants of CTT have a small run-time overhead, and can be customised with the equivalent of cryptographic keys. The stronger variants of CTT have extra training but this is used to characterise propagation bounds rather than to defend against specific adversarial samples, yielding a more flexible and general defense mechanism.

\clearpage

%% file: sections/appendix.tex
\appendix
\label{apdx:act}
\appendixpage
\addappheadtotoc

\section{Training procedure}
In this section we explain how to train a CTT instrumented
model.
First, most of the commonly used optimisers are suitable for
CTT training. However, it should be noted that there
exists an interaction between the CTT penalty
(the additional loss term introduced by CTT) and the weight decay of the optimizer.
Although we have not evaluated this interaction formally, we find it easier to train models when the weight decay is
either turned off or set to a very small value.
The optimizer used in our experiments is RMSProp.

The annealing procedure for CTT parameters is important for convergence.
The parameter $\alpha$ determines the strength of the CTT penalty, and
we increase $\alpha$ iteratively by a factor of $\beta$ every
$t$ training epochs.
For both MNIST and FashionMNIST, we used $t=6, \beta=0.005$.
For CIFAR10, we used $t=30, \beta=0.001$. We find that the best way to train the models is to first optimise
$L_{V}$, i.e. make sure that neurons have a bound larger than $T_{l} $ and then start iteratively increasing $\alpha$. We hypothesise that this works in line with recent findings that there exist a number of connected convergence clusters with similar performance~\cite{fort2019deep} with a path between them. Iteratively increasing $\alpha$ allows us to keep convergence, while maintaining low $L_{V}$ loss and decreasing the false positive rate.

\section{Parameter Selection}

In this section we try to explain intuition behind CTT and the
instrumentation parameter ($\epsilon$) choice.
Algorithm 1 in the paper presents the whole procedure
we have used to successfully train
both CTT-loose and CTT-strict.

First, the parameter
$\epsilon$ can be thought of as a \textit{detectability}
certification of an adversarial sample.
It defines the minimum theoretical perturbation size for which the detector should work.
In other words, when training the classifier with CTT, we generate
adversarial bounds up to a limit of $\epsilon$.
Rather than generating adversarial
samples, we use a natural sample perturbed by $\epsilon$.
The CTT loss tries to ensure that when adversarial samples
$X\pm\epsilon$ are considered, the detector neurons can be
turned on.

CTT may be understood in contrast with the work of Cohen et al. in certifiable
robustness ~\cite{cohen2019certified}.
Certifiable
robustness aims at making natural sample behaviour stay in a pre-formed
$l_p$ ball, so that model behaviour is stable in a natural range of values.
CTT, on the other hand, aims at detecting illegal behaviours outside of this $l_p$ ball,
so that behaviour outside of the natural range is unstable.
The intuition is that the smaller you make this $l_p$ norm for CTT, the
the easier we can detect adversarial behaviours.

\begin{figure}[!tbp]
  \centering
  \includegraphics[width=0.7\linewidth]{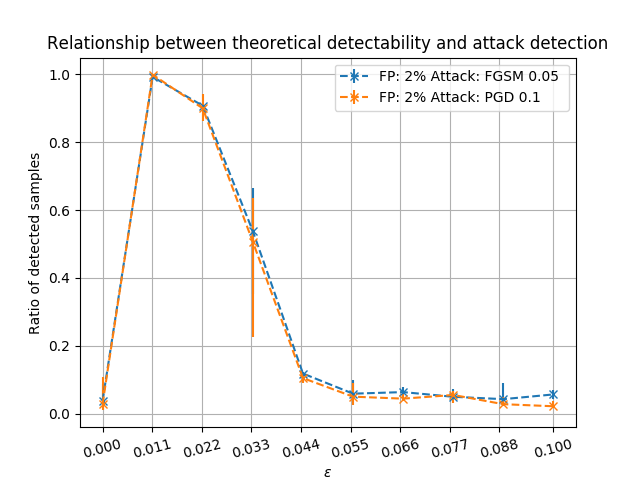}
  \caption{
    Trade-off between choices of $\epsilon$ and detector performance.
    There are five
    LeNet5 networks classifying
    MNIST, instrumented with CTT-loose with $T_{l} = 10^{-4}$
    with a given $\epsilon$.
    The networks are trained to a false positive rate of $2\%$.
    Points show median performance,
    whereas error bars refer to standard deviations of the $5$ networks.}
  \label{fig:epsilonp}
\end{figure}

\Cref{fig:epsilonp} shows the trade-off between detector
performance and $\epsilon$-choice using CTT-loose.
First, the pre-defined $\epsilon$ does not guarantee that
all adversarial samples above this $\epsilon$ will be detected
(only CTT-strict guarantees this).
Second, there is an optimal $\epsilon$ we can choose
to maximize the detectability.
Third, unlike adversarial training, CTT does not rely on a
particular attack during training.
As shown in \Cref{fig:epsilonp},
the performance of different attacks (PGD and FGSM)
shows almost no difference and the attack quality seems to be only related to the
perturbation size rather than generation procedure.
Finally, the left-most point shown in \Cref{fig:epsilonp} refers to $\epsilon=10^{-5}$.
It seems that for a given architecture and a given dataset there
exists a smallest $\epsilon$ one could successfully train the model
for. For MNIST with LeNet5 we struggled to find a reproducible way to
train the model for epsilons below $0.0007$.

\section{FashionMNIST vs MNIST}

MNIST is a popular benchmark, but is known to be
relatively simple~\cite{lecun2010mnist}. Xiao et al. proposed
FashionMNIST~\cite{xiao2017online}, a more complex, yet still simple toy dataset.
In this section we report on results of CTT-loose instrumentation of
LeNet5 networks solving FashionMNIST with $\epsilon=0.001$,
meaning that the adversarial set includes perturbed images
with an $l_\infty$ size of $0.001$.

In addition to the attacks presented in the evaluation section,
we also show here the results for
a decision-based attack~\cite{brendel2018decisionbased}. 
The attack itself is particularly interesting as it 
is not based on any gradient information, meaning CTT detection is not network-information specific. 
For this attack, we
use 25 trials per iteration and vary the number of iterations.

\Cref{tab:fashion_mnist} shows the results of attacking CTT-loose
instrumented LeNet5.
In the evaluation section of the paper we have shown that for
MNIST CTT detection was capable of capturing almost all of the adversarial samples.
Unlike MNIST, CTT fails to detect all of the adversarial samples on FashionMNIST.

As already noted,
there is a relationship between the attack perturbation size, dataset specifics, and the detectability of CTT.
In the case of FashionMNIST, for the particular $\epsilon$ value, we find it shows
relatively better detection rate for small $l_2$ values.


\input{tables/fashion.tex}

\section{Detectability trade-off}

\begin{figure}[!tbp]
  \centering
  \includegraphics[width=0.7\linewidth]{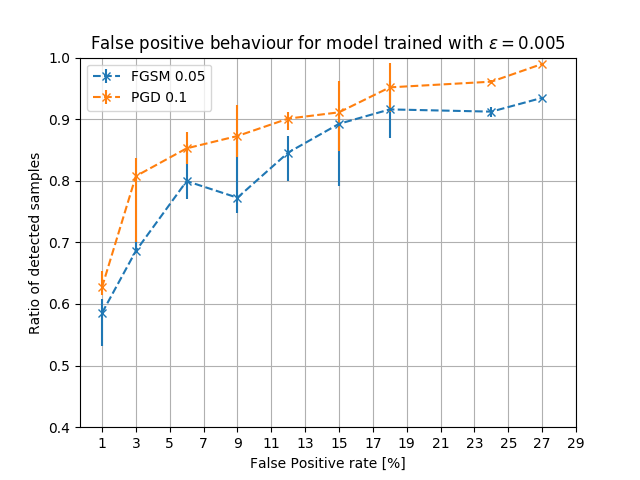}
  \caption{Trade-off between choices of false positive rates and detector performance.
  There are five LeNet5 networks classifying FashionMNIST,
  instrumented with CTT-loose with $T_{l} = 10^{-4}$ with a given $\epsilon=0.005$.
  Points show median performance,
  whereas error bars refer to standard deviations.}
  \label{fig:fp_performance}
\end{figure}

In this section we show the impact of different
false positive rates on the CTT-loose instrumentation.
We use 5 LeNet5 networks and train each of them with
the same CTT-loose restrictions but stop at different
training times so that
networks achieve different false-positive rates.
\Cref{fig:fp_performance} presents the false positive rate trade-off two specific attacks, with false
positive rates on the x-axis
and detectability on the y-axis.

The relationship between detector performance and
false positive rates indicates a trade-off of interest when applying CTT-loose in practice.
With a slight increase of false positive rates ($1\%$ to $3\%$),
we increase the detector performance by around $20\%$.
Intuitively,
this suggests first, that there exist inefficiencies in the
internal representations of the neural network,
where the network struggles to separate natural and non-natural samples.
Second, this trade-off between false positive rates
and detectability can occur because of imperfections of the training
dataset.
The natural training dataset involves imperfect, confusing images
for the network, and thus causing a vague boundary between the
natural and adversarial input sets.
It should be noted that
although this relationship exists across different datasets and models,
its scaling seems to be dataset-dependent.

\section{Run-time overheads}
Previously, \citeauthor{shumailov2018taboo}
demonstrated that Taboo Trap adds no extra
inference cost and utilises zero additional device memory when deployed.
In comparison, MagNet \cite{meng2017magnet}
and SafetyNet \cite{lu2017safetynet}
show $20\%$ and $3600\%$ increases in terms
of additional parameters when deployed \cite{shumailov2018taboo}.
As mentioned in our paper, CTT
shares the same detection mindset as Taboo Trap,
and thus enjoy the benefit of having zero extra cost in network inference as well.
Unlike \cite{shumailov2018taboo} however we instrument less than 0.1\% of all neurons
available in the network. That in turn means that fewer detector neurons are required for much better detection.

%% file: tables/fashion.tex
\begin{table}[!h]
\centering
\caption{
 CTT-loose instrumented LeNet5 network classifying FashionMNIST.
}

\begin{adjustbox}{scale=0.8,center}
\begin{tabular}{@{}llll|ll@{}}
\toprule
& $\theta$ & $l_{2}$ & \multicolumn{2}{c}{CTT-loose} & \\
\midrule
No Attack       & && $85.3\%\ (5.3\%)$ & $84.4\%\ (2.7\%)$\\
\midrule
\multirow{6}{*}{FGSM} & $\epsilon = 0.006$ & 0.10 & $64.73\%\ (75.33\%)$ & $65.14\%\ (39.66\%)$ & \\
                      & $\epsilon = 0.007$ & 0.12 & $60.79\%\ (71.01\%)$ & $62.74\%\ (39.68\%)$ &\\
                      & $\epsilon = 0.01$ & 0.17 & $52.32\%\ (61.80\%)$ & $56.97\%\ (38.55\%)$&\\
                      & $\epsilon = 0.03$ & 0.52 & $22.62\%\ (54.42\%)$ & $24.40\%\ (44.52\%)$&\\
                      & $\epsilon = 0.05$ & 0.88 & $11.72\%\ (56.11\%)$ & $12.62\%\ (49.79\%)$&\\
                      & $\epsilon = 0.07$ & 1.23 & $6.61\%\ (59.63\%)$ & $8.29\%\ (51.51\%)$&\\
 \midrule
\multirow{6}{*}{Boundary} & $i = 10$ & 1.96 &$0.00\%\ (28.19\%)$ & $0.00\%\ (20.19\%)$ &\\
								 & $i=50$   & 1.62 &$0.00\%\ (27.61\%)$ & $0.00\%\ (23.44\%)$ &\\
								 & $i=100$  & 1.23 &$0.00\%\ (29.23\%)$ & $0.00\%\ (24.52\%)$&\\
								 & $i=500$ &  0.24 &$0.00\%\ (60.90\%)$ & $0.00\%\ (38.94\%)$&\\
								 & $i=1000$ & 0.15 & $0.00\%\ (65.31\%)$ & $0.00\%\ (44.27\%)$&\\
\bottomrule
\end{tabular}
\end{adjustbox}

\label{tab:fashion_mnist}
\end{table}

%% file: paper.bbl
\begin{thebibliography}{43}
\providecommand{\natexlab}[1]{#1}
\providecommand{\url}[1]{\texttt{#1}}
\expandafter\ifx\csname urlstyle\endcsname\relax
  \providecommand{\doi}[1]{doi: #1}\else
  \providecommand{\doi}{doi: \begingroup \urlstyle{rm}\Url}\fi

\bibitem[Badrinarayanan et~al.(2015)Badrinarayanan, Kendall, and
  Cipolla]{badrinarayanan2015segnet}
Badrinarayanan, V., Kendall, A., and Cipolla, R.
\newblock Segnet: A deep convolutional encoder-decoder architecture for image
  segmentation.
\newblock \emph{arXiv preprint arXiv:1511.00561}, 2015.

\bibitem[Bhagoji et~al.(2018)Bhagoji, He, Li, and Song]{bhagoji2018black}
Bhagoji, A.~N., He, W., Li, B., and Song, D.
\newblock Black-box attacks on deep neural networks via gradient estimation.
\newblock \emph{International Conference on Learning Representations Workshop
  (ICLR)}, 2018.

\bibitem[Brendel et~al.(2018)Brendel, Rauber, and
  Bethge]{brendel2018decisionbased}
Brendel, W., Rauber, J., and Bethge, M.
\newblock Decision-based adversarial attacks: Reliable attacks against
  black-box machine learning models.
\newblock In \emph{International Conference on Learning Representations}, 2018.
\newblock URL \url{https://openreview.net/forum?id=SyZI0GWCZ}.

\bibitem[Carlini \& Wagner(2017{\natexlab{a}})Carlini and
  Wagner]{carlini2017magnet}
Carlini, N. and Wagner, D.
\newblock Magnet and" efficient defenses against adversarial attacks" are not
  robust to adversarial examples.
\newblock \emph{arXiv preprint arXiv:1711.08478}, 2017{\natexlab{a}}.

\bibitem[Carlini \& Wagner(2017{\natexlab{b}})Carlini and
  Wagner]{carlini2017towards}
Carlini, N. and Wagner, D.
\newblock {Towards Evaluating the Robustness of Neural Networks}.
\newblock In \emph{2017 IEEE Symposium on Security and Privacy (SP)}, pp.\
  39--57. IEEE, 2017{\natexlab{b}}.

\bibitem[Carlini et~al.(2016)Carlini, Mishra, Vaidya, Zhang, Sherr, Shields,
  Wagner, and Zhou]{Carlini2016}
Carlini, N., Mishra, P., Vaidya, T., Zhang, Y., Sherr, M., Shields, C., Wagner,
  D., and Zhou, W.
\newblock {H}idden {V}oice {C}ommands.
\newblock In \emph{25th {USENIX} Security Symposium ({USENIX} Security 16)}.
  {USENIX} Association, 2016.

\bibitem[Chen et~al.(2019)Chen, Carlini, and Wagner]{chen2019stateful}
Chen, S., Carlini, N., and Wagner, D.~A.
\newblock Stateful detection of black-box adversarial attacks.
\newblock \emph{CoRR}, abs/1907.05587, 2019.
\newblock URL \url{http://arxiv.org/abs/1907.05587}.

\bibitem[Cohen et~al.(2019)Cohen, Rosenfeld, and Kolter]{cohen2019certified}
Cohen, J., Rosenfeld, E., and Kolter, Z.
\newblock Certified adversarial robustness via randomized smoothing.
\newblock In \emph{Proceedings of the 36th International Conference on Machine
  Learning}, 2019.

\bibitem[Dvijotham et~al.(2018)Dvijotham, Gowal, Stanforth, Arandjelovic,
  O'Donoghue, Uesato, and Kohli]{dvijotham2018training}
Dvijotham, K., Gowal, S., Stanforth, R., Arandjelovic, R., O'Donoghue, B.,
  Uesato, J., and Kohli, P.
\newblock Training verified learners with learned verifiers.
\newblock \emph{arXiv preprint arXiv:1805.10265}, 2018.

\bibitem[Eykholt et~al.(2018)Eykholt, Evtimov, Fernandes, Li, Rahmati, Xiao,
  Prakash, Kohno, and Song]{eykholt2018robust}
Eykholt, K., Evtimov, I., Fernandes, E., Li, B., Rahmati, A., Xiao, C.,
  Prakash, A., Kohno, T., and Song, D.
\newblock {R}obust {P}hysical-{W}orld {A}ttacks on {D}eep {L}earning {V}isual
  {C}lassification.
\newblock In \emph{Proceedings of the IEEE Conference on Computer Vision and
  Pattern Recognition}, pp.\  1625--1634, 2018.

\bibitem[Fang et~al.(2003)Fang, Chen, and Fuh]{fang2003road}
Fang, C.-Y., Chen, S.-W., and Fuh, C.-S.
\newblock Road-sign detection and tracking.
\newblock \emph{IEEE transactions on vehicular technology}, 52\penalty0
  (5):\penalty0 1329--1341, 2003.

\bibitem[Fort et~al.(2019)Fort, Hu, and Lakshminarayanan]{fort2019deep}
Fort, S., Hu, H., and Lakshminarayanan, B.
\newblock Deep ensembles: A loss landscape perspective.
\newblock \emph{arXiv preprint arXiv:1912.02757}, 2019.

\bibitem[Gehr et~al.(2018)Gehr, Mirman, Drachsler-Cohen, Tsankov, Chaudhuri,
  and Vechev]{gehr2018ai2}
Gehr, T., Mirman, M., Drachsler-Cohen, D., Tsankov, P., Chaudhuri, S., and
  Vechev, M.
\newblock Ai2: Safety and robustness certification of neural networks with
  abstract interpretation.
\newblock In \emph{2018 IEEE Symposium on Security and Privacy (SP)}, pp.\
  3--18. IEEE, 2018.

\bibitem[Goodfellow et~al.(2015)Goodfellow, Shlens, and
  Szegedy]{goodfellow2014explaining}
Goodfellow, I.~J., Shlens, J., and Szegedy, C.
\newblock Explaining and harnessing adversarial examples.
\newblock \emph{International Conference on Learning Representations (ICLR)},
  2015.

\bibitem[Gowal et~al.(2018)Gowal, Dvijotham, Stanforth, Bunel, Qin, Uesato,
  Mann, and Kohli]{gowal2018effectiveness}
Gowal, S., Dvijotham, K., Stanforth, R., Bunel, R., Qin, C., Uesato, J., Mann,
  T., and Kohli, P.
\newblock On the effectiveness of interval bound propagation for training
  verifiably robust models.
\newblock \emph{arXiv preprint arXiv:1810.12715}, 2018.

\bibitem[Ji et~al.(2013)Ji, Xu, Yang, and Yu]{ji20133d}
Ji, S., Xu, W., Yang, M., and Yu, K.
\newblock 3d convolutional neural networks for human action recognition.
\newblock \emph{IEEE transactions on pattern analysis and machine
  intelligence}, 35\penalty0 (1):\penalty0 221--231, 2013.

\bibitem[Katz et~al.(2017)Katz, Barrett, Dill, Julian, and
  Kochenderfer]{katz2017reluplex}
Katz, G., Barrett, C., Dill, D.~L., Julian, K., and Kochenderfer, M.~J.
\newblock Reluplex: An efficient smt solver for verifying deep neural networks.
\newblock In \emph{International Conference on Computer Aided Verification},
  pp.\  97--117. Springer, 2017.

\bibitem[Krizhevsky et~al.(2012)Krizhevsky, Sutskever, and
  Hinton]{krizhevsky2012imagenet}
Krizhevsky, A., Sutskever, I., and Hinton, G.~E.
\newblock Image{N}et classification with deep convolutional neural networks.
\newblock In \emph{Advances in neural information processing systems}, pp.\
  1097--1105, 2012.

\bibitem[Krizhevsky et~al.(2014)Krizhevsky, Nair, and
  Hinton]{krizhevsky2014cifar}
Krizhevsky, A., Nair, V., and Hinton, G.
\newblock The {CIFAR-10} dataset.
\newblock 2014.

\bibitem[Kurakin et~al.(2017)Kurakin, Goodfellow, and
  Bengio]{kurakin2016adversarial}
Kurakin, A., Goodfellow, I., and Bengio, S.
\newblock Adversarial machine learning at scale.
\newblock 2017.

\bibitem[LeCun et~al.(2010)LeCun, Cortes, and Burges]{lecun2010mnist}
LeCun, Y., Cortes, C., and Burges, C.
\newblock {MNIST} handwritten digit database.
\newblock 2, 2010.

\bibitem[LeCun et~al.(2015)]{lecun2015lenet}
LeCun, Y. et~al.
\newblock {LeNet-5}, convolutional neural networks.
\newblock pp.\ ~20, 2015.

\bibitem[L{\'e}cuyer et~al.(2018)L{\'e}cuyer, Atlidakis, Geambasu, Hsu, and
  Jana]{lcuyer2018certified}
L{\'e}cuyer, M., Atlidakis, V., Geambasu, R., Hsu, D., and Jana, S.
\newblock Certified robustness to adversarial examples with differential
  privacy.
\newblock In \emph{IEEE S\&P 2019}, 2018.

\bibitem[Lu et~al.()Lu, Issaranon, and Forsyth]{lu2017safetynet}
Lu, J., Issaranon, T., and Forsyth, D.~A.
\newblock Safetynet: Detecting and rejecting adversarial examples robustly.

\bibitem[Madry et~al.(2018)Madry, Makelov, Schmidt, Tsipras, and
  Vladu]{madry2017towards}
Madry, A., Makelov, A., Schmidt, L., Tsipras, D., and Vladu, A.
\newblock Towards deep learning models resistant to adversarial attacks.
\newblock 2018.

\bibitem[Meng \& Chen(2017)Meng and Chen]{meng2017magnet}
Meng, D. and Chen, H.
\newblock Magnet: A two-pronged defense against adversarial examples.
\newblock In \emph{Proceedings of the 2017 ACM SIGSAC Conference on Computer
  and Communications Security}, CCS '17, pp.\  135--147, New York, NY, USA,
  2017. ACM.

\bibitem[Metzen et~al.(2017)Metzen, Genewein, Fischer, and
  Bischoff]{metzen2017detecting}
Metzen, J.~H., Genewein, T., Fischer, V., and Bischoff, B.
\newblock On detecting adversarial perturbations.
\newblock In \emph{Proceedings of 5th International Conference on Learning
  Representations (ICLR)}, 2017.

\bibitem[Mirman et~al.(2018)Mirman, Gehr, and Vechev]{mirman2018differentiable}
Mirman, M., Gehr, T., and Vechev, M.
\newblock Differentiable abstract interpretation for provably robust neural
  networks.
\newblock In \emph{International Conference on Machine Learning}, 2018.

\bibitem[Mustafa et~al.(2019)Mustafa, Khan, Hayat, Goecke, Shen, and
  Shao]{mustafa2019adversarial}
Mustafa, A., Khan, S., Hayat, M., Goecke, R., Shen, J., and Shao, L.
\newblock Adversarial defense by restricting the hidden space of deep neural
  networks.
\newblock In \emph{The IEEE International Conference on Computer Vision
  (ICCV)}, October 2019.

\bibitem[Pang et~al.(2019)Pang, Xu, Du, Chen, and Zhu]{pang2019improving}
Pang, T., Xu, K., Du, C., Chen, N., and Zhu, J.
\newblock Improving adversarial robustness via promoting ensemble diversity.
\newblock In Chaudhuri, K. and Salakhutdinov, R. (eds.), \emph{Proceedings of
  the 36th International Conference on Machine Learning}, volume~97 of
  \emph{Proceedings of Machine Learning Research}, pp.\  4970--4979, 2019.

\bibitem[Rauber et~al.(2017)Rauber, Brendel, and Bethge]{rauber2017foolbox}
Rauber, J., Brendel, W., and Bethge, M.
\newblock Foolbox: A python toolbox to benchmark the robustness of machine
  learning models.
\newblock \emph{arXiv preprint arXiv:1707.04131}, 2017.

\bibitem[Ren et~al.(2015)Ren, He, Girshick, and Sun]{ren2015faster}
Ren, S., He, K., Girshick, R., and Sun, J.
\newblock Faster {R-CNN}: Towards real-time object detection with region
  proposal networks.
\newblock In \emph{Advances in neural information processing systems}, pp.\
  91--99, 2015.

\bibitem[Schott et~al.(2019)Schott, Rauber, Bethge, and
  Brendel]{schott2018towards}
Schott, L., Rauber, J., Bethge, M., and Brendel, W.
\newblock Towards the first adversarially robust neural network model on mnist.
\newblock \emph{International Conference on Learning Representations Workshop
  (ICLR)}, 2019.

\bibitem[Schroff et~al.(2015)Schroff, Kalenichenko, and
  Philbin]{schroff2015facenet}
Schroff, F., Kalenichenko, D., and Philbin, J.
\newblock Facenet: A unified embedding for face recognition and clustering.
\newblock In \emph{Proceedings of the IEEE conference on computer vision and
  pattern recognition}, pp.\  815--823, 2015.

\bibitem[Shan et~al.(2019)Shan, Willson, Wang, Li, Zheng, and
  Zhao]{shan2019gotta}
Shan, S., Willson, E., Wang, B., Li, B., Zheng, H., and Zhao, B.~Y.
\newblock Gotta catch 'em all: Using concealed trapdoors to detect adversarial
  attacks on neural networks.
\newblock \emph{CoRR}, abs/1904.08554, 2019.
\newblock URL \url{http://arxiv.org/abs/1904.08554}.

\bibitem[Shumailov et~al.(2018)Shumailov, Zhao, Mullins, and
  Anderson]{shumailov2018taboo}
Shumailov, I., Zhao, Y., Mullins, R., and Anderson, R.
\newblock The taboo trap: Behavioural detection of adversarial samples.
\newblock \emph{arXiv preprint arXiv:1811.07375}, 2018.

\bibitem[Shumailov et~al.(2019)Shumailov, Gao, Zhao, Mullins, Anderson, and
  Xu]{shumailov2019sitatapatra}
Shumailov, I., Gao, X., Zhao, Y., Mullins, R., Anderson, R., and Xu, C.-Z.
\newblock Sitatapatra: Blocking the transfer of adversarial samples.
\newblock 2019.

\bibitem[Szegedy et~al.(2013)Szegedy, Zaremba, Sutskever, Bruna, Erhan,
  Goodfellow, and Fergus]{Szegedy2013}
Szegedy, C., Zaremba, W., Sutskever, I., Bruna, J., Erhan, D., Goodfellow,
  I.~J., and Fergus, R.
\newblock Intriguing properties of neural networks.
\newblock \emph{CoRR}, abs/1312.6199, 2013.

\bibitem[Wang et~al.(2018)Wang, Pei, Whitehouse, Yang, and
  Jana]{wang2018efficient}
Wang, S., Pei, K., Whitehouse, J., Yang, J., and Jana, S.
\newblock Efficient formal safety analysis of neural networks.
\newblock In \emph{Advances in Neural Information Processing Systems}, pp.\
  6367--6377, 2018.

\bibitem[Wong et~al.(2018)Wong, Schmidt, Metzen, and Kolter]{wong2018scaling}
Wong, E., Schmidt, F., Metzen, J.~H., and Kolter, J.~Z.
\newblock Scaling provable adversarial defenses.
\newblock In \emph{Advances in Neural Information Processing Systems}, pp.\
  8400--8409, 2018.

\bibitem[Xiao et~al.(2017)Xiao, Rasul, and Vollgraf]{xiao2017online}
Xiao, H., Rasul, K., and Vollgraf, R.
\newblock Fashion-mnist: a novel image dataset for benchmarking machine
  learning algorithms.
\newblock 2017.

\bibitem[Zhao et~al.(2018{\natexlab{a}})Zhao, Gao, Mullins, and
  Xu]{zhao2018mayo}
Zhao, Y., Gao, X., Mullins, R., and Xu, C.
\newblock Mayo: A framework for auto-generating hardware friendly deep neural
  networks.
\newblock 2018{\natexlab{a}}.

\bibitem[Zhao et~al.(2018{\natexlab{b}})Zhao, Shumailov, Mullins, and
  Anderson]{zhao2018compress}
Zhao, Y., Shumailov, I., Mullins, R., and Anderson, R.
\newblock To compress or not to compress: Understanding the interactions
  between adversarial attacks and neural network compression.
\newblock \emph{arXiv preprint arXiv:1810.00208}, 2018{\natexlab{b}}.

\end{thebibliography}
